\documentclass[sigconf, nonacm, 9pt]{acmart}
\AtBeginDocument{%
  }

\setcopyright{acmlicensed}
\copyrightyear{2026}
\acmYear{2026}
\acmDOI{XXXXXXX.XXXXXXX}
\acmConference[MLCAD '26]{ACM/IEEE International Symposium on Machine Learning for CAD}{September 7--9,
  2026}{Jeju Island, South Korea}

\usepackage{booktabs}
\usepackage{multirow}
\usepackage{graphicx}
\usepackage{caption}
\usepackage{subcaption}
\usepackage{longtable}
\usepackage{hhline}

\usepackage{enumitem}
\usepackage{microtype}
\usepackage{graphicx}
\usepackage{booktabs}
\usepackage{tikz}
\usepackage{amsmath}
\usepackage{mathtools}
\usepackage{subcaption}
\usepackage{amsthm}
\usepackage{textcomp}
\usepackage{amsmath}

\usepackage[capitalize,noabbrev]{cleveref}

\theoremstyle{plain}

\theoremstyle{definition}

\theoremstyle{remark}

\usepackage[textsize=tiny]{todonotes}
\usepackage{multirow}
\usepackage{graphicx}
\usepackage{caption}
\usepackage{longtable}

\begin{document}

\title{Differentiable Initialization-Accelerated CPU-GPU Hybrid Combinatorial Scheduling}

\author{Mingju Liu}
\email{mliu9867@umd.edu}
\affiliation{%
  \institution{University of Maryland, College Park}
  \state{Maryland}
  \country{USA}
}

\author{Jiaqi Yin}
\email{jyin629@umd.edu}
\affiliation{%
  \institution{University of Maryland, College Park}
  \state{Maryland}
  \country{USA}
}

\author{Alvaro Velasquez}
\email{alvaro.velasquez@colorado.edu}
\affiliation{%
  \institution{University of Colorado, Boulder}
  \state{Colorado}
  \country{USA}
}

\author{Cunxi Yu}
\email{cunxiyu@umd.edu}
\affiliation{%
  \institution{University of Maryland, College Park}
  \state{Maryland}
  \country{USA}
}

\begin{abstract}
This paper presents a hybrid CPU-GPU framework for solving combinatorial scheduling problems formulated as Integer Linear Programming (ILP). While scheduling underpins many optimization tasks in computing systems, solving these problems optimally at scale remains a long-standing challenge due to their NP-hard nature. We introduce a novel approach that combines differentiable optimization with classical ILP solving. Specifically, we utilize differentiable presolving to rapidly generate high-quality partial solutions, which serve as warm-starts for commercial ILP solvers (CPLEX, Gurobi) and rising open-source solver HiGHS. This method enables significantly improved early pruning compared to state-of-the-art standalone solvers. Empirical results across industry-scale benchmarks demonstrate up to a $10\times$ performance gain over baselines, narrowing the optimality gap to $<0.1\%$. This work represents the first demonstration of utilizing differentiable optimization to initialize exact ILP solvers for combinatorial scheduling, opening new opportunities to integrate machine learning infrastructure with classical exact optimization methods across broader domains.
\end{abstract}





\maketitle

\section{Introduction}
\label{sec:introduction}

Combinatorial scheduling lies at the core of many optimization problems in computing systems, where the goal is to assign tasks over time and resources under dependency and capacity constraints. It plays a pivotal role in domains such as hardware synthesis, where efficient scheduling directly impacts performance, area, and power consumption. However, the problem is \textit{NP-hard}, and designing scalable and optimal solutions remains a long-standing challenge.

One of the most successful formulations of scheduling problems uses \textit{System of Difference Constraints (SDC)}, a subclass of the more general \textit{Integer Linear Programming (ILP)} systems. In these systems, each linear constraint involves at most two variables (e.g., $x_i - x_j \leq b$), which compactly captures task dependencies and timing/resource constraints \cite{cong2006efficient,dai2018scalable}. These constraints can be directly encoded in ILP, enabling the derivation of provably optimal schedules. This method has been widely adopted in commercial and industrial tools such as AMD Xilinx Vivado/Vitis HLS \cite{kathail2020xilinx,cong2011high} and Google's XLS \cite{babu2021xls}. 

Despite the advantage of optimality, ILP-based methods suffer from limited scalability due to exponential computational complexity and reliance on general-purpose solvers. To address this, {heuristic algorithms} such as \cite{paulin1989force,blum2003metaheuristics,ahn2020ordering} have been developed to provide fast, often problem-specific solutions that sacrifice optimality for runtime efficiency.

More recently, \textbf{machine learning (ML)} techniques are explored to overcome limitations of exact and heuristic methods by leveraging high-performance computing platforms like GPUs \cite{sanders2010cuda} and TPUs \cite{jouppi2017datacenter}. These techniques broadly fall into three categories:

\begin{itemize}[leftmargin=1em]
    \item \textbf{Imitation Learning (IL)} \cite{baltean2018strong, gasse2019exact, gagrani2022neural, wang2023cardinality, han2023gnn}: These approaches learn policies from expert demonstrations or optimal solutions. While effective in constrained scenarios, they struggle with generalization due to limited data diversity and domain shift.
    
    \item \textbf{Reinforcement Learning (RL)} \cite{mascia2014grammar, karapetyan2017markov, chen2019RLS, yin2023respect, yin2023accelerating, yu2020flowtune, neto2022end}: RL learns directly from interaction and reward signals, enabling the discovery of novel scheduling strategies. However, it often suffers from large runtime overhead and unstable convergence in large-scale problems.
    
    \item \textbf{Differentiable Optimization} The recent state-of-the-art (SOTA) differentiable optimization framework \cite{liu2024diffsched} has emerged that forms a combinatorial scheduling problem using stochastic relaxation and gradient-based optimization. Built upon the Gumbel-Softmax \cite{jang2016categorical}, this method enables training-free, dataless optimization with customizable objectives. While it demonstrates fast convergence and competitive results on large benchmarks, it still yields sub-optimal schedules due to non-convexity and approximation inherent in the relaxation.
\end{itemize}

In this work, we propose a novel combinatorial scheduling framework that integrates SDC/ILP formulations with differentiable optimization to provide a hybrid CPU-GPU solution balancing both \textit{optimality} and \textit{scalability}. Specifically, we introduce a two-stage hybrid scheduling flow that leverages differentiable optimization to address the sub-optimality of pure gradient-based methods, while simultaneously accelerating ILP solving. 

The \textbf{key novelty} of this work lies in using differentiable optimization—via gradient-descent algorithms—as a \textit{differentiable warm-start} for exact ILP solvers, which aims to bring a seamless integration between machine learning infrastructure into exact ILP solving. Default ILP pre-solving techniques often rely on heuristics or prior feasible solutions, which are difficult to generate for complex scheduling instances due to significant runtime consumption and offer limited coverage of the solution space. As demonstrated in our experimental results compared to SOTA commercial tools with all warm-start options enabled, our method uses a differentiable relaxation based on the constrained Gumbel Trick to rapidly explore feasible regions and generate high-quality partial solutions. These solutions are then used to warm-start SOTA ILP solvers, enabling faster convergence to optimal or near-optimal schedules. To the best of our knowledge, this is the first approach that leverages differentiable optimization as an initialization mechanism to accelerate exact ILP solvers for combinatorial scheduling problems and offers new opportunities in CPU-GPU hybrid combinatorial optimization.

Our methodology, built on SDC constraints, comprises: (1) fast differentiable optimization with on-the-fly, confidence-based partial solution selection, and (2) warm-starting commercial and open-source SOTA ILP solvers such as CPLEX \cite{cplex2009v12}, Gurobi \cite{gurobi}, and HiGHS \cite{huangfu2018parallelizing}. This hybrid flow achieves significantly improved scheduling performance by combining the efficiency of CPU-GPU accelerated differentiable solvers with the robustness of ILP-based exact methods. Our experiment results demonstrate significant improvements over SOTAs, while more importantly maintaining the optimality and determinism guarantees. This differentiable warm-start two-stage hybrid optimization workflow provides the potential to be extended beyond combinatorial scheduling to general ILP problems. Our code and setups will be open-sourced.

\section{Preliminary}
\label{sec:preliminary}

\subsection{Scheduling and Problem statement}
Scheduling is a well-established combinatorial challenge with extensive real-world relevance. In this work, we schedule tasks in a dataflow graph modeled as a directed acyclic graph (DAG) \(G(V,E)\), where each node \(v \in V\) represents a computational task and each edge \(e \in E\) encodes a data dependency between graph nodes. 
In many settings, cost metrics are assigned to nodes and/or edges in the form of weights. 
Additionally, various constraints—such as timing requirements or limitations on available resources—may be imposed depending on the specific scheduling scenario. 
We define a schedule \(\mathbf{S} = (s_0, s_1, \ldots, s_i)\), where \(s_i\) denotes the stage in which node \(v_i\) is executed. Under a given latency \(L\), representing the maximum completion time of the entire DAG, the objective is to minimize resource usage (and/or other specified costs) while respecting all dependency constraints. This formulation is referred to as a \emph{latency-constrained min-resource scheduling} problem.

Various approaches have been explored to address combinatorial scheduling challenges in compiler optimization and hardware synthesis. Machine learning (ML) techniques, particularly Topoformer~\cite{gagrani2022neural}, introduce attention-based graph neural networks for topological ordering through node embedding learning. \cite{han2023gnn} combines ML with optimization and proposes a novel predict-and-search framework for efficiently identifying high-quality feasible solutions. While promising, such supervised learning approaches face limitations in generalizability and data dependency. Reinforcement learning (RL) with graph learning-based schedulers~\cite{chen2019RLS,yin2023respect,yin2023accelerating} attempts to address these limitations by learning from action rewards, though challenges persist in scalability and training overhead.


\subsection{SDC and Related Methods}

\begin{figure}[htb]   
  \centering
  \includegraphics[width=0.95\columnwidth]{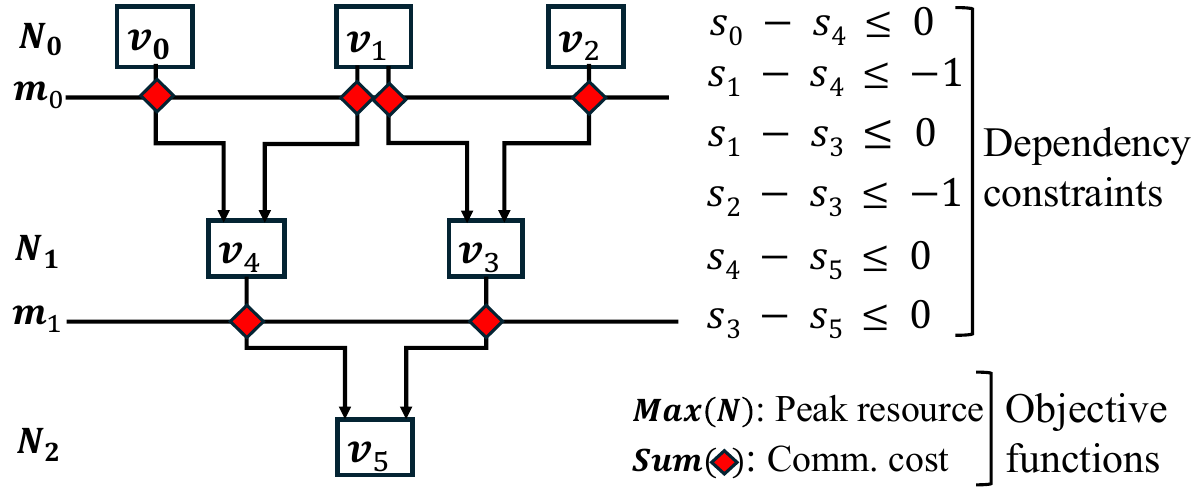}
  \caption{Example of SDC-based scheduling. (Left) A DFG with three scheduling stages, having a latency of $L=3$. (Right) The corresponding dependency constraints and objective functions include minimizing peak resource and inter-stage communication costs.}
  \vspace{-0.3cm}
  \label{fig:sdc-example}
\end{figure}
\textbf{System of Difference Constraints (SDC)}, which consists of inequalities of the form \(x_i - x_j \leq c_{ij}\) with integer constants \(c_{ij}\), is a prominent example of ILP systems. SDC formulations enable efficient polynomial-time solutions while ensuring integer feasibility, making them valuable for scheduling applications. 
We demonstrate the SDC-based scheduling formulation using a simple data flow graph (DFG) illustrated in Figure~\ref{fig:sdc-example}. The formulation handles operation dependencies by introducing inequality constraints for each edge in the DFG. Specifically, for a data edge connecting operation $i$ to operation $j$, SDC imposes the constraint $s_i - s_j \leq c$, where $s_i$ and $s_j$ represent the scheduled stage of operations $i$ and $j$, respectively. For instance, given the edge from node $v_0$ to node $v_4$ in our example, the formulation includes the constraint $s_0 - s_4 \leq 0$, ensuring that operation $v_4$ is scheduled at or after the stage of operation $v_0$. This pattern extends to all data-dependent edges in the DFG. 



\section{Approach}
\label{sec:approach}

The key novelty of our proposed approach lies in integrating differentiable optimization via gradient-descent algorithms directly with ILP solving, while simultaneously enabling CPU-GPU co-optimization to solve combinatorial scheduling problems at scale while pursuing optimality.
Specifically, our solving scheme consists of a high-performance two-stage hybrid optimization flow that enables (1) differentiable scheduling and confidence-guided warm-start generation (GPU/CPU), and (2) integrated multi-thread parallel ILP solving utilizing the generated warm-starts (CPU).

\subsection{Warm-start With a Partial Solution Set}\label{sec:partial}
Similar to classic presolving techniques widely used in SOTA ILP solvers, providing a high-quality initial solution aims to guide solvers toward promising regions of the solution space more efficiently—particularly for large-scale problems.
These warm-start strategies are often applied during preprocessing or used for optimality gap analysis.
The differentiable scheduling method proposed in \cite{liu2024diffsched} offers consistently fast convergence and yields reasonably good solutions within a short timeframe, especially compared to traditional ILP solvers.

Modern ILP solvers like CPLEX, Gurobi, and HiGHS allow users to provide initial additional solutions through commands like \textit{setStart}, \textit{v.Start}, or \textit{setSolution}.
These variable assignments can guide the branch-and-bound search, helping solvers identify promising regions and avoid suboptimal branches.
This capability is crucial for large-scale or time-limited problems, where the exponential worst-case complexity of ILP formulations makes exhaustive search impractical.
High-quality solutions serving as "warm-starts" can significantly reduce solving time by constraining the effective search space.
Note that these warm-starts do not fix variables; instead, they serve as non-binding hints. The solver is not forced to adhere to these values and remains free to explore the entire solution space.
Consequently, the original problem formulation is never altered, and the solver's ability to prove optimality is fully preserved.

\begin{figure}[t]   
  \centering
  \includegraphics[width=\columnwidth]{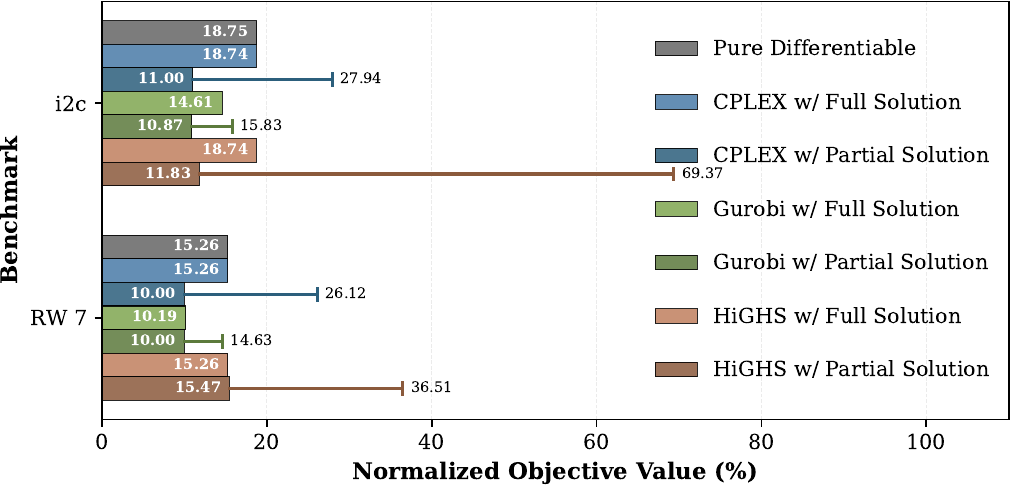}
  \vspace{-0.6cm}
  \caption{Comparison of warm-start strategies: full solution vs. partial solution.
  Normalized objective values are obtained at a 900s solving time.
  Error bars represent the range of objective values obtained when warm-starting different solvers with varying partial solutions.}
  \label{fig:motiv}
  \vspace{-2em}
\end{figure}

However, the final output of such a differentiable method is often sub-optimal and may become trapped in local minima once convergence is reached.
To address this, and before introducing our proposed differentiable warm-start generation strategy, we conduct a detailed case study based on the following three \textbf{Hypotheses}: 
\textbf{(1)} The default pre-solving techniques of SOTA ILP solvers can be significantly improved;
\textbf{(2)} The full solution produced by differentiable methods may not serve as a good warm-start for ILP solvers due to local optima;
and \textbf{(3)} despite this, these local optima often contain high-quality partial solutions—i.e., good initializations over subsets of variables—which can be extracted to create effective warm-starts for exact ILP solving.

We present a case study to empirically demonstrate these key hypotheses.
We select two benchmarks from the suites used in \cite{liu2024diffsched}: \texttt{i2c} (small-scale) and RW 7 (large-scale).
For each benchmark, we perform 30 iterations of optimization using the pure differentiable method.
The converged results are recorded, and the solution for each schedule variable is used as a \textit{full solution} warm-start for CPLEX, Gurobi, and HiGHS.
Additionally, we sample 30 partial solution sets from the full solution to serve as \textit{partial solution} warm-starts.
All ILP solver experiments are conducted with a 900s timeout. The final objective values are presented in Figure~\ref{fig:motiv}.
Note that the objective value is normalized by the heuristic solution (see Section \ref{sec:experiment}) of each benchmark and is thus presented as a percentage.

In Figure~\ref{fig:motiv}, we observe two key findings:
(1) SOTA ILP solvers frequently encounter challenges when using a full solution as a warm-start.
For instance, both CPLEX and HiGHS stagnated at initialization (showing no improvement over the pure differentiable baseline) until the timeout when solving \texttt{i2c} and RW7 with a full solution.
Gurobi performs slightly better, continuing to optimize, but still stagnates at sub-optimal values.
(2) Partial solution warm-starts offer greater potential for reaching near-optimal solutions.
By leveraging sampled partial solutions of varying sizes, most ILP solvers demonstrate improved performance across both benchmarks.
However, it is worth noting that certain partial solutions can lead to worse objective values.
For instance, HiGHS solves \texttt{i2c} to an objective value range of $11.83\%$ to $69.37\%$ (indicated by the error bars in Figure~\ref{fig:motiv}), emphasizing the importance of developing a robust strategy for selecting high-performance warm-starts.

\subsection{Confidence-Based Differentiable Partial Solution Generation} \label{sssec:confidence}

Following the observations from the case study, we leverage a confidence-based strategy to construct high-quality partial solutions.
In general, for a DAG $G(V,E)$, let $\mathbf{S} = \{s_1, s_2, ..., s_i, ..., s_{|V|}\}$ represent the complete set of schedule variable solutions.
Each variable $s_i$ is associated with a confidence value $\mathbf{C}_i \in [0,1]$ that quantifies the certainty of its assignment.
This confidence estimation enables us to extract the subset of variables most likely to contribute to the optimal solution.
Formally, we define our partial solution extraction mechanism as follows:

Given the confidence set $\mathbf{C} = \{\mathbf{C}_i\}_{i=1}^{|V|}$ for a full solution, we select variables for the partial solution warm-start using a preset threshold:
\begin{equation}
    \mathcal{F} = \{ s_i \mid \mathbf{C}_i \geq \tau_{\mathbf{C}} \}, \quad \tau_{\mathbf{C}} \in [0,1]
    \label{eq:confidence}
\end{equation}
where $\tau_{\mathbf{C}}$ controls solution completeness; specifically, higher $\tau_{\mathbf{C}}$ values yield smaller partial solutions for initializing SOTA ILP solvers.
We construct these confidence values using the differentiable method in \cite{liu2024diffsched}, which leverages the gradient information of each variable to provide a principled estimate of solution certainty within our hybrid optimization workflow.

\subsubsection{Differentiable SDC formulation}\label{sec:ilp2}
\cite{liu2024diffsched} formulates the latency-constrained min-resource scheduling problem to schedule a node set $V$ on $L$ pre-defined stages, minimizing cost objectives while allowing node chaining.
By utilizing SDC, node dependencies are translated into integer linear inequalities, ensuring the solution schedule retains data flow integrity.
Specifically, the inequalities are formulated with respect to the edge set $E$:
\begin{equation}  
    \forall e(i,j) \in E: s_i - s_j \leq c_{ij}
    \label{eq:sdc_k_2_t}
\end{equation}
where \( e(i,j) \) denotes an edge connecting node \( i \) to node \( j \), and \( s_i \) and \( s_j \) are the schedule variables for nodes \( i \) and \( j \), respectively.
Given a latency constraint \( L \), any schedule variable must satisfy $s_i \leq L - 1$ (0-indexed). 

To fully encode this scheduling problem as a differentiable formulation, two core transformations are required:

\paragraph{Search Space Vectorization} 
The core insight from prior work lies in representing discrete scheduling decisions as continuous probability vectors.
For each variable $s_i$ within latency bound $L$, we construct a one-hot vector $\overrightarrow{s_i}$ via hard Gumbel-Softmax on the probability vector $\overrightarrow{P_i}$:
\begin{equation}
    \overrightarrow{s_i} = \text{hard GS}(\overrightarrow{P_i}), \quad \overrightarrow{P_i} \in \mathbb{R}^L \label{eq:hard}
\end{equation}
where $\text{argmax}(\overrightarrow{s_i}) \in [0, L-1]$ bijectively maps to integer scheduling stages.
This vectorization transforms the entire search space into a $\mathbb{R}^{|V|\times L}$ tensor for DAG $G(V,E)$, enabling efficient gradient computations.
For example, scheduling a node at stage $t=2$ in a 3-stage system ($L=3$) corresponds to the vector $[0,0,1]$.

\paragraph{Differentiable Constraint Encoding} 
We use the \textit{constrained Gumbel Trick} to restrict the feasible solution space for dependent variables through element-wise multiplication during Gumbel sampling:
\begin{equation}
    y'_i = \frac{\exp((\log p_i + g_i)/\tau)}{\sum_j \exp((\log p_j + g_j)/\tau)} \odot \mathbf{T_{\leq}}(\overrightarrow{s_i}, c_{ij}) \label{eq:con_gs}
\end{equation}
where $\mathbf{T_{\leq}}(\overrightarrow{s_i}, c_{ij}) = \texttt{cumsum}(\overrightarrow{s_i} \hat{+} |c_{ij}|)$, and $\hat{+}$ denotes a right-shift operator on the one-hot vector.
As a concrete example, consider $s_i$ encoded as $[0,1,0]$ with $c_{ij}=0$: the resulting $\mathbf{T_{\leq}} = [0,1,1]$ forces $s_j$ to be scheduled no earlier than stage 1, maintaining differentiable optimization.
This mechanism ensures all dependency constraints are respected throughout the optimization process.

\subsubsection{Differentiable Optimization} \label{ssec:opt}

\paragraph{Constrained Probability Formulation} 
The optimization process employs constrained Gumbel Trick sampling to generate valid schedules.
Using vectorized representations $\overrightarrow{s}$, we calculate sampling probabilities by obtaining a probability vector $\overrightarrow{P}$:
\begin{equation}
    \overrightarrow{P} = \text{GS}(\mathbf{T_{\leq}}(\overrightarrow{s}, c))\label{eq:prob_vector}
\end{equation}
where \( \overrightarrow{P} \in \mathbb{R}^L \) and \( |\overrightarrow{P}^a| \in [0, 1] \) for each stage \( a \in [0, L-1] \), and $\mathbf{T_{\leq}}$ enforces encoded inequality constraints.
Each probability vector serves as the determinant for the final stage assignment of the current node and is updated iteratively by gradient descent during optimization.

\paragraph{Differentiable Cost Objectives} 
We minimize two objectives jointly as stated in \cite{liu2024diffsched}:

\textbf{(1) Peak Memory Resource}: Modeled through entropy loss ~\cite{wang2010multi}:
\begin{equation}
    \mathcal{L}_{r} = -\sum_{a=0}^{L-1} \frac{N_{a}}{|V|}\log \frac{N_{a}}{|V|}
\end{equation}
where $N_a$ is the number of nodes on stage $a$.

\textbf{(2) Communication Cost}: Calculated as:
\begin{equation}
    \mathcal{L}_{c} = \frac{1}{\sum w_b} \sum_{a=0}^{L-2} m_a
\end{equation}
where $w_b$ is the communication cost introduced by an edge $e_b \in E$, and $m_a$ is the inter-stage communication cost (between stage $a$ and $a+1$).

The composite cost is $\mathcal{L} = \lambda\mathcal{L}_{r} + \mathcal{L}_{c}$, where $\lambda$ is a customizable ratio parameter adjusting the trade-off between the two objectives.

\paragraph{Gradient-guided Differentiable Warm-start}
Our method selects high-quality partial solutions for warm-starting using the confidence metrics from Equation \ref{eq:confidence}.
Specifically, the confidence value $\mathbf{C}_i$ is derived from the constrained probability distributions.
For each schedule variable $s_i$ with constrained probability vector $\overrightarrow{P'_i} \in [0,1]^L$ (Eq.~\ref{eq:prob_vector}), we define:
\begin{equation}
    \mathbf{C}_i = \max(\overrightarrow{P'_i}) \in [0,1] \label{eq:confi}
\end{equation}

\textbf{Example}: Given a current probability vector $\overrightarrow{P_i} = [0.5,0.2,0.3]$ and constraint $\mathbf{T_{\leq}} = [0,1,1]$, the probabilities become $\overrightarrow{P'_i} = [0,0.2,0.3]$ with $\mathbf{C}_i = 0.3$ at stage 2. If the pre-defined $\tau_{\mathbf{C}} = 0.3$, then this variable will be selected for the warm-start.

\subsection{Two-Stage Hybrid CPU-GPU Optimization Workflow} \label{ssec:2step}

This section details our complete workflow, combining differentiable initialization with SOTA ILP solvers.
The workflow operates through two sequential phases:

\paragraph{\textbf{Stage 1: Differentiable Warm-Start Generation (CPU+GPU)}}  
The process begins with vectorized initialization of schedule variables in $\mathbb{R}^{|V|\times L}$ as defined in Section~\ref{sec:ilp2}.
Through iterative constrained Gumbel sampling (Eq.~\ref{eq:con_gs}), probability vectors are updated while tracking confidence metrics $\mathbf{C}_i$ per variable (Eq.~\ref{eq:confi}).
At each iteration $k \in [1, t]$, partial solutions $\mathcal{F}_k$ are extracted by thresholding confidence values: $\mathcal{F}_k = \{ s_i \mid \mathbf{C}_i^{(k)} \geq \tau_{\mathbf{C}} \}$, where $\tau_\mathbf{C}$ can be customized and controls solution completeness.
This generates $t$ distinct warm-starts, representing high-confidence scheduling decisions. Note that differentiable solving is allocated as a GPU workload to enable massive parallel initial solving, while the final partial warm-start solution generation occurs on the CPU.

\paragraph{\textbf{Stage 2: Parallelized Solver Execution (CPU)}}  
Each warm-start $\mathcal{F}_k$ initializes SOTA ILP solvers by assigning variables $x_i = \hat{x}_i \ \forall i \in \mathcal{F}_k$. The solvers then optimize the same SDC ILP scheduling problem formulation, leveraging generated warm-starts to accelerate the solving process.
After parallel execution across all $t$ warm-starts, the optimal solution $\mathcal{J}^* = \min\{\mathcal{J}_1, ..., \mathcal{J}_t\}$ is selected, combining the speed of the gradient-based differentiable method with the optimality guarantees of exact SOTA ILP solvers.
\section{Experiment}
\label{sec:experiment}

\begin{table}[h!]
\centering
\caption{Terminology Definitions}
\label{tab:terminology}
\begin{tabular}{@{}ll@{}}
\toprule
\textbf{Term} & \textbf{Definition} \\
\midrule
\textbf{DiffSched} & The differentiable method introduced in \cite{liu2024diffsched}. \\
\textbf{Cold-start} & Solver execution without additional initial solutions. \\
\textbf{Warm-start} & Solver execution initialized with additional hints. \\
\bottomrule
\end{tabular}
\vspace{-1mm}
\end{table}

\paragraph{\textbf{Benchmarks}}
We evaluate our approach on combinatorial scheduling problems derived from 6 designs in the EPFL Benchmark Suite \cite{amaru2015epfl}, extended by incorporating GPU computational graphs of post-mapped designs using the 7nm ASAP library \cite{xu2017standard}. Additionally, we utilize 12 synthetic random workloads (RW) of varying sizes and densities. Details of all 24 benchmarks are summarized in Table~\ref{tab:diff_conv}.

\begin{table}[h!]
\vspace{-0.8em}
\caption{Benchmark Details. \#Constraints: ILP formulation constraint count; T/iter: \texttt{DiffSched} runtime per iteration; Conv.: converged objective value of \texttt{DiffSched}.}
\label{tab:diff_conv}
\resizebox{\columnwidth}{!}{%
\begin{tabular}{|c|c|c|c||c|c|c|c|}
\hhline{|----||----|}
\textbf{Design} & \textbf{\#Constraints} & \textbf{T/iter (s)} & \textbf{Conv.} & \textbf{Design} & \textbf{\#Constraints} & \textbf{T/iter (s)} & \textbf{Conv.} \\ 
\hhline{|====||====|}
\textbf{Adder} & 93066 & 5.38 & 21.03\% & \textbf{RW 1} & 102927 & 3.36 & 17.78\% \\ \hhline{|-|-|-|-||-|-|-|-|}
\textbf{Adder (M)} & 122930 & 6.68 & 22.96\% & \textbf{RW 2} & 105131 & 3.43 & 19.86\% \\ \hhline{|-|-|-|-||-|-|-|-|}
\textbf{Bar} & 264082 & 14.73 & 20.07\% & \textbf{RW 3} & 104071 & 5.07 & 16.72\% \\ \hhline{|-|-|-|-||-|-|-|-|}
\textbf{Bar (M)} & 239166 & 11.07 & 20.60\% & \textbf{RW 4} & 36641 & 1.22 & 16.31\% \\ \hhline{|-|-|-|-||-|-|-|-|}
\textbf{i2c} & 114124 & 6.51 & 18.75\% & \textbf{RW 5} & 38213 & 1.27 & 16.94\% \\ \hhline{|-|-|-|-||-|-|-|-|}
\textbf{i2c (M)} & 96182 & 5.01 & 17.63\% & \textbf{RW 6} & 41267 & 2.60 & 16.98\% \\ \hhline{|-|-|-|-||-|-|-|-|}
\textbf{Max} & 243821 & 13.58 & 37.97\% & \textbf{RW 7} & 196264 & 7.10 & 15.26\% \\ \hhline{|-|-|-|-||-|-|-|-|}
\textbf{Max (M)} & 292127 & 13.87 & 23.55\% & \textbf{RW 8} & 192146 & 6.94 & 16.32\% \\ \hhline{|-|-|-|-||-|-|-|-|}
\textbf{Square} & 1412460 & 56.07 & 23.39\% & \textbf{RW 9} & 194005 & 12.80 & 16.76\% \\ \hhline{|-|-|-|-||-|-|-|-|}
\textbf{Square (M)} & 1854792 & 62.80 & 22.78\% & \textbf{RW 10} & 378920 & 13.75 & 14.90\% \\ \hhline{|-|-|-|-||-|-|-|-|}
\textbf{Voter} & 1086767 & 41.63 & 18.45\% & \textbf{RW 11} & 389326 & 18.82 & 15.39\% \\ \hhline{|-|-|-|-||-|-|-|-|}
\textbf{Voter (M)} & 2107732 & 63.26 & 14.88\% & \textbf{RW 12} & 387700 & 23.78 & 14.77\% \\ \hhline{|----||----|}
\end{tabular}%
}
\vspace{-1em}
\end{table}

\paragraph{\textbf{Baselines and Solution Set Collection}}
For fair comparison, we adopt hyperparameters from \cite{liu2024diffsched}: the entropy-communication loss ratio $\lambda$ is 100 for EPFL designs and 10 for RWs. We use a latency of $L=10$ for all problems. Experiments are conducted on an Intel\textregistered Xeon\textregistered Gold 6418H CPU and an NVIDIA RTX\texttrademark A6000 GPU, with a 3600s timeout. We run 30 differentiable iterations per design, collecting 30 solution sets as warm-starts for CPLEX \cite{cplex2009v12}, Gurobi \cite{gurobi}, and HiGHS \cite{huangfu2018parallelizing}. 

The confidence thresholds are set to $\tau_{\mathbf{C}} = 0.2$ (RW and post-mapped EPFL) and $\tau_{\mathbf{C}} = 0.127$ (original EPFL). We normalize the solved objective value using heuristic values from CPLEX. Unless specified otherwise, “Objective Value” always refers to this normalized metric, and “Performance” always refers to the solved objective value at the timeout. In Table \ref{tab:diff_conv}, \texttt{DiffSched} converged objective values are recorded at the 30th iteration. As noted in \cite{liu2024diffsched}, \texttt{DiffSched} typically converges within the first 20 iterations, ensuring the recorded values reflect true converged performance. Both \texttt{DiffSched} results in Table \ref{tab:diff_conv} and cold-start solver results at 3600s serve as baselines.

\paragraph{\textbf{Solvers Setup}} We utilize the Python APIs of three SOTA ILP solvers (CPLEX, Gurobi, HiGHS) to optimize the formulated ILP problems. To ensure fairness, solvers' built-in parallel solving is disabled (i.e., single-thread only). For Cold-start baseline, we initiate solvers with 30 distinct random seeds running in parallel to mimic our hybrid Warm-start method's behavior, which warm-starts with 30 different solution sets in parallel. We report the result from the seed producing the best (lowest) objective value as the Cold-start solver result -- aligns with the Warm-start result extraction.

\subsection{Performance Comparison}\label{sec:comp}
We structure the experimental results to answer the following research questions (RQ):

\noindent
\textbf{\underline{RQ1}: How should confidence thresholds be set, and is the method sensitive to threshold selection?}

\begin{figure}[t]
    \centering
    \begin{subfigure}[b]{0.48\columnwidth}
        \centering
        \includegraphics[width=\linewidth]{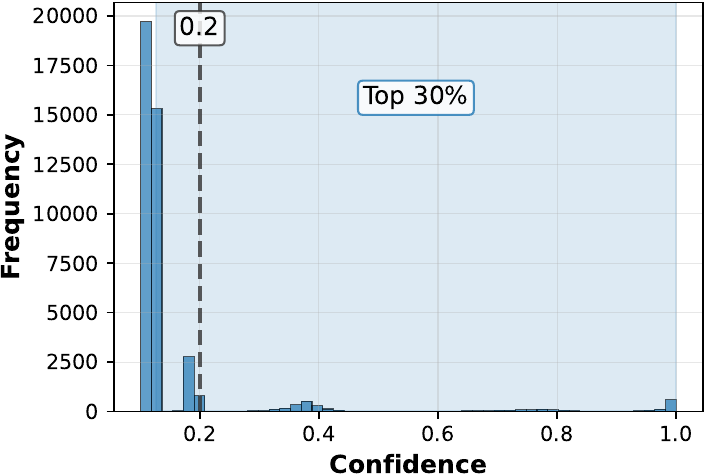}
        \caption{Distribution: RW designs}
        \label{fig:hist_rw}
    \end{subfigure}
    \hfill
    \begin{subfigure}[b]{0.48\columnwidth}
        \centering
        \includegraphics[width=\linewidth]{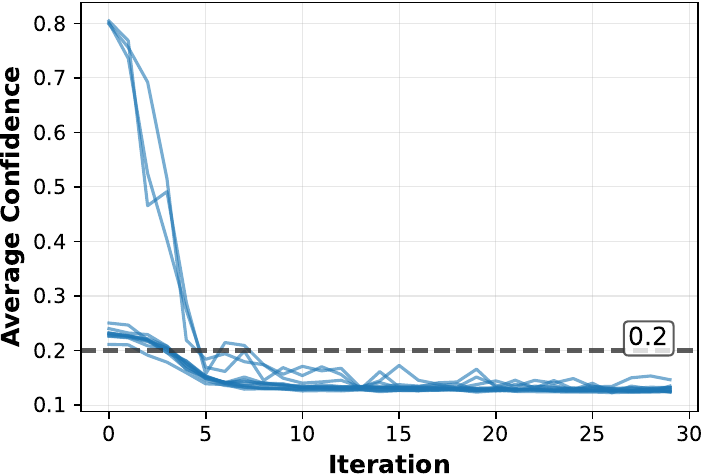}
        \caption{Iteration: RW designs}
        \label{fig:iter_rw}
    \end{subfigure}
    \begin{subfigure}[b]{0.48\columnwidth}
        \centering
        \includegraphics[width=\linewidth]{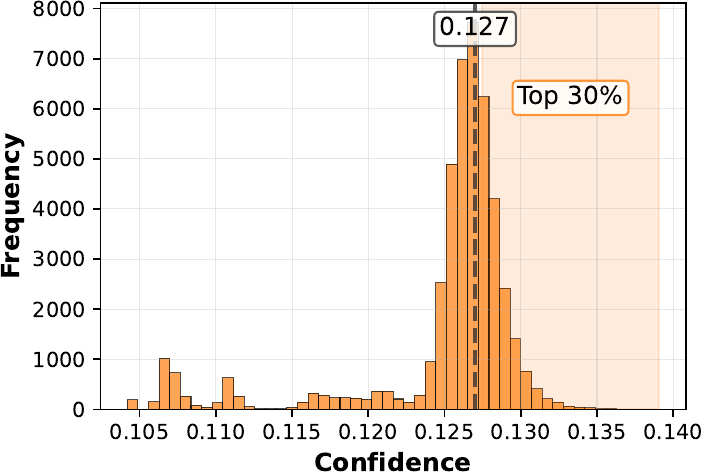}
        \caption{Distribution: EPFL designs}
        \label{fig:hist_el}
    \end{subfigure}
    \hfill
    \begin{subfigure}[b]{0.48\columnwidth}
        \centering
        \includegraphics[width=\linewidth]{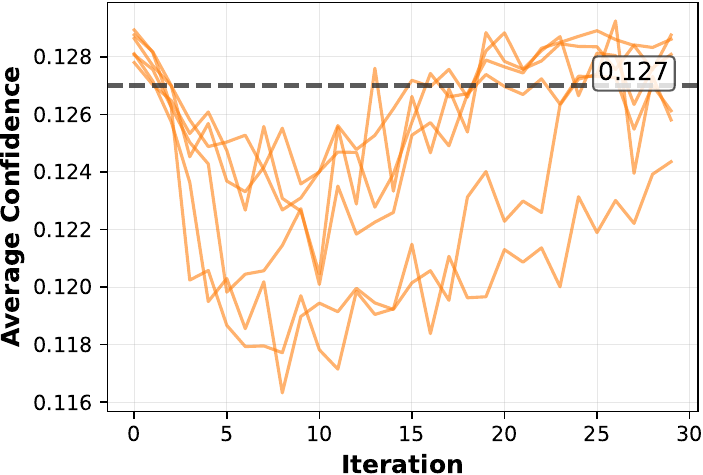}
        \caption{Iteration: EPFL designs}
        \label{fig:iter_el}
    \end{subfigure}
    \begin{subfigure}[b]{0.48\columnwidth}
        \centering
        \includegraphics[width=\linewidth]{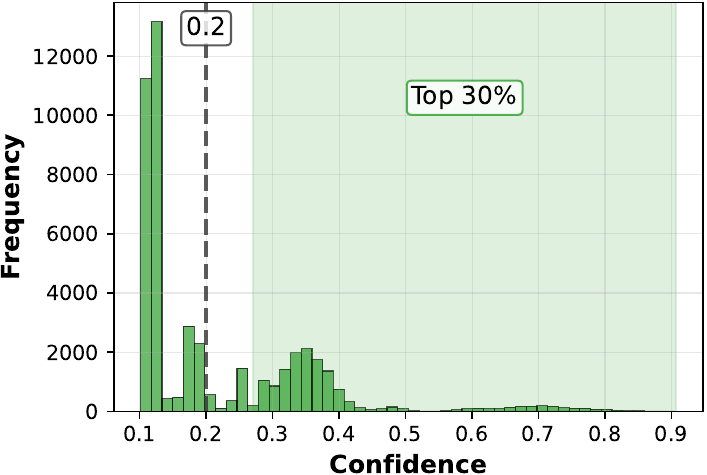}
        \caption{Distribution: EPFL(M) designs}
        \label{fig:hist_el_mapped}
    \end{subfigure}
    \hfill
    \begin{subfigure}[b]{0.48\columnwidth}
        \centering
        \includegraphics[width=\linewidth]{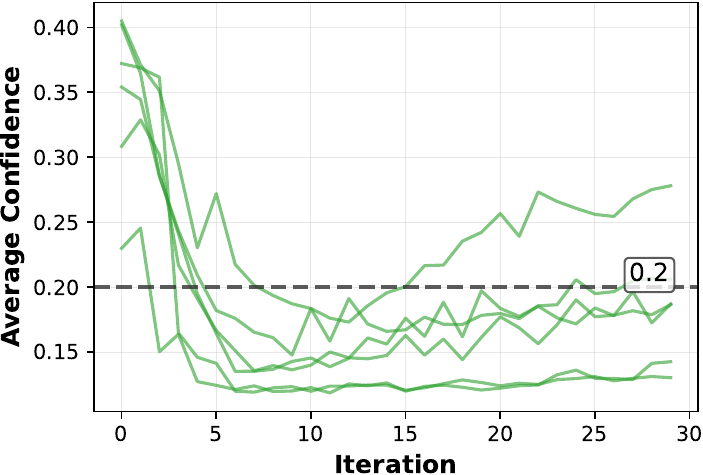}
        \caption{Iteration: EPFL(M) designs}
        \label{fig:iter_el_mapped}
    \end{subfigure}
    \caption{Confidence analysis: Left: Aggregated confidence distributions with the top 30\% region shaded. Right: Average confidence for the top 30\% assignments across 30 iterations. Rows correspond to RW, EPFL, and mapped EPFL designs.}
    \label{fig:threshold_analysis}
    \vspace{-2em}
\end{figure}

\textbf{Thresholds are benchmark-specific but show low sensitivity within reasonable ranges.} As discussed in Section~\ref{sec:partial}, partial solutions are effective warm-starts when they initialize sufficient variables to establish viable starting points without over-specification. Following this, we typically select $5\%$ to $15\%$ of total variables.
Figures~\ref{fig:hist_rw}, \ref{fig:hist_el}, and \ref{fig:hist_el_mapped} display confidence distributions aggregated across all solutions of all benchmarks in each benchmark category. The shaded regions highlight the top 30\% most confident assignments used for guiding partial solution selection.
And Figures~\ref{fig:iter_rw}, \ref{fig:iter_el}, and \ref{fig:iter_el_mapped} track the average confidence of these top 30\% assignments across 30 iterations. 

Although confidence values are aggregated per category, individual solutions vary. For example, Figure \ref{fig:hist_el} shows that the top 30\% assignments have a threshold slightly below 0.127, while Figure \ref{fig:iter_el} shows that between iterations 5 and 15, the average confidence drops below 0.127. \textbf{In our workflow setup}, we perform an automated rapid post-processing step based on the confidence aggregation to guide and customize the selection of the confidence threshold. This analysis indicates that $\tau_{\mathbf{C}}$ can be adjusted within a reasonable range without significantly degrading solution quality, as the relative ordering of high-confidence assignments remains stable across iterations.

\begin{figure*}[t]
  \centering
  \includegraphics[width=\textwidth]{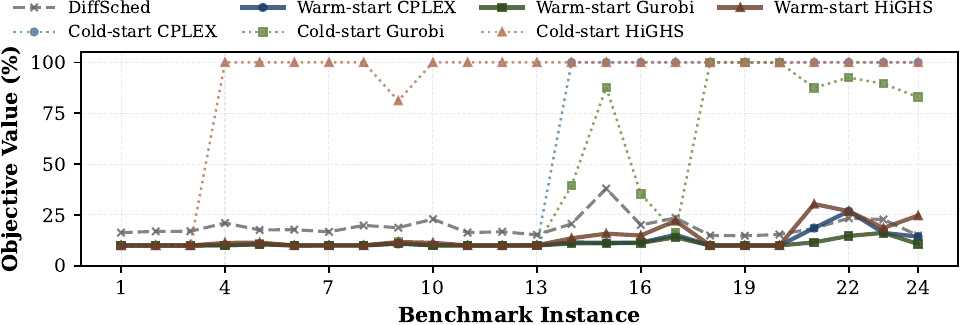}
  \caption{Performance comparison across 24 benchmark instances, sorted by ILP constraint count (small to large). Cold-start/Warm-start results are collected at 3600s/1800s, respectively. Lower objective value indicates better performance.}
  \label{fig:result_comparison}
\end{figure*}

\begin{table*}[t]
\centering
\caption{Average Performance and Improvements Across Benchmark Categories}
\label{tab:bench_results_detailed}
\resizebox{\textwidth}{!}{%
\begin{tabular}{lccccccccccccc}
\toprule
\multirow{2}{*}{\textbf{Category}} & \multirow{2}{*}{\texttt{\textbf{DiffSched}}\cite{liu2024diffsched}} & \multicolumn{4}{c}{\textbf{CPLEX}\cite{cplex2009v12}} & \multicolumn{4}{c}{\textbf{Gurobi}\cite{gurobi}} & \multicolumn{4}{c}{\textbf{HiGHS}\cite{huangfu2018parallelizing}} \\
\cmidrule(lr){3-6} \cmidrule(lr){7-10} \cmidrule(lr){11-14}
& & Cold-start & Warm-start & $\Delta_1$ & $\Delta_2$ & Cold-start & Warm-start & $\Delta_1$ & $\Delta_2$ & Cold-start & Warm-start & $\Delta_1$ & $\Delta_2$ \\
\midrule
EPFL & 21.84\% & 70.13\% & \textbf{13.96\%} & -7.88\% & -56.17\% & 47.82\% & \textbf{11.84\%} & -10.00\% & -35.97\% & 98.43\% & \textbf{17.72\%} & -4.12\% & -80.71\% \\
RW & 16.50\% & 32.51\% & \textbf{10.01\%} & -6.49\% & -22.50\% & 32.50\% & \textbf{10.00\%} & -6.50\% & -22.49\% & 77.50\% & \textbf{10.03\%} & -6.47\% & -67.47\% \\
Total & 19.17\% & 51.32\% & \textbf{11.99\%} & -7.18\% & -39.33\% & 40.16\% & \textbf{10.92\%} & -8.25\% & -29.23\% & 87.96\% & \textbf{13.87\%} & -5.30\% & -74.09\% \\
\midrule
\multicolumn{14}{l}{\footnotesize Results collected at 3600s (Cold-start) and 1800s (Warm-start). $\Delta_1$ = Warm-start - \texttt{DiffSched}; $\Delta_2$ = Warm-start - Cold-start. Negative values indicate improvement.} \\
\bottomrule
\end{tabular}%
}
\end{table*}

\noindent
\textbf{\underline {RQ2}: What is the overhead of collecting partial solutions from \texttt{DiffSched}?}

\textbf{The overhead for collection is manageable.} Table \ref{tab:diff_conv} shows iteration runtime varies by benchmark size. Small instances (RW 4, RW 5) require approximately 1 second per iteration, while the largest design (\texttt{Voter(M)}) requires roughly 1 minute. Since we collect 30 solution sets over 30 iterations, the total collection overhead ranges from 30 seconds to 30 minutes. This overhead also highly depends on computational resources and can be reduced by deploying \texttt{DiffSched} on newer GPU systems.

\noindent
\textbf{\underline {RQ3}: Do ILP solvers with differentiable warm-start achieve better trade-offs between solution quality and runtime?}

\textbf{ILP solvers with differentiable warm-starts achieve better trade-offs, particularly on larger benchmarks.} 
To account for the collection overhead mentioned in \textbf{\underline{RQ2}}, we compare Cold-start results at 3600s against Warm-start results at 1800s, which incorporates the upper bound for collection time as 1800s, and we present these results in Figure \ref{fig:result_comparison} and Table \ref{tab:bench_results_detailed}. Note that the optimal objective value for this problem formulation can be viewed as $10.00 \pm 0.01\%$.

In Figure \ref{fig:result_comparison}, benchmark instances are sorted by ILP formulation constraint count (from small to large)\footnote{RW4, RW5, RW6, \texttt{Adder}, \texttt{i2c(M)}, RW1, RW3, RW2, \texttt{i2c}, \texttt{Adder(M)}, RW8, RW9, RW7, \texttt{Bar(M)}, \texttt{Max}, \texttt{Bar}, \texttt{Max(M)}, RW10, RW12, RW11, \texttt{Voter}, \texttt{Square}, \texttt{Square(M)}, \texttt{Voter(M)}.} to highlight performance vs. scalability. Both Cold-start commercial solvers perform well on smaller benchmarks (instances 1-13), but suffer performance degradation after instance 14. Meanwhile, open-source solver HiGHS shows severe scalability issues starting at instance 4. 

With our provided differentiable warm-starts, performance significantly improves for all solvers, further bridging the barrier between open-source and commercial solvers. For example, instance 15 (\texttt{Max}), the Cold-start solver's objective value stuck around 87\%, while \texttt{DiffSched} can reach 37.97\%; however, Warm-start method improves the objective to 11\% for the two commercial solvers and 16\% for open-source solver HiGHS. Similarly, for instance 19 (RW 12), all Cold-start solvers fail to initialize (staying at 100\% at timeout), whereas all Warm-start solvers achieve near-optimal 10\% values, indicating up to 10x performance improvement.

Table \ref{tab:bench_results_detailed} reinforces these improvements by displaying exact numbers of the average performance across different benchmark categories. Note that RW designs are generally easier than EPFL designs due to their nature of synthetic structure. Cold-start HiGHS averages 98.43\% on EPFL designs which is compared to 70.13\% for Cold-start CPLEX and 47.82\% for Cold-start Gurobi. With warm-starts, HiGHS improves to 17.72\%, within 6\% of the best-performing Warm-start Gurobi, which is less than 2\% near-optimal. Overall, Warm-start method shows over 5\% performance improvement when compared to \texttt{DiffSched}, and over 39\% improvement compared to their Cold-start implementation. It indicates that differentiable warm-starts can enhance the ability of commercial SOTA solvers to reach optimal within a limited timeframe and enable open-source solvers to compete with commercial tools.

\noindent
\textbf{\underline {RQ4}: Does the order of \texttt{DiffSched} solutions impact the proposed Warm-start method's performance?}

\begin{figure}[t]
  \centering
  \begin{subfigure}[b]{0.48\columnwidth}
    \includegraphics[width=\linewidth]{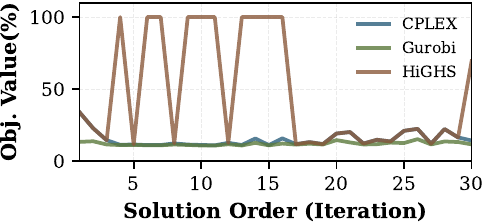}
    \caption{i2c}
    \label{fig:conv_i2c}
  \end{subfigure}
  \hfill
  \begin{subfigure}[b]{0.48\columnwidth}
    \includegraphics[width=\linewidth]{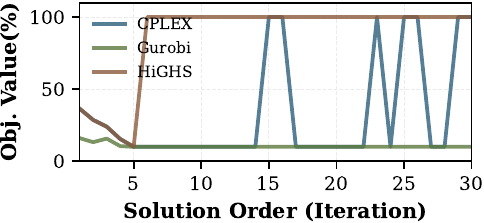}
    \caption{RW 7}
    \label{fig:conv_rw7}
  \end{subfigure}
  \vspace{-1em}
  \caption{Solver performance vs. partial solution order.}
  \label{fig:convergence}
\end{figure}

\textbf{The solution order has negligible impact on the final parallelized solving performance.} Our setup utilizes parallel execution for both Cold-start and Warm-start implementation, recording only the best result at runtime, hence the solution order can be neglected when deciding the result. To analyze the impact of specific solution orders obtained at different \texttt{DiffSched} iterations, we use Figure \ref{fig:convergence} to compare solver performance on \texttt{i2c} and RW 7. 
We can observe that different solvers react differently even to the same initial solution. In Figure \ref{fig:conv_i2c}, HiGHS exhibits high volatility, alternating between successful and failed optimization depending on the specific iteration. Similarly, Figure \ref{fig:conv_rw7} shows that while Gurobi remains robust, CPLEX fails to improve upon solutions from specific iterations (e.g., iteration 15). Although individual solutions vary in quality, our parallelization strategy mitigates this issue by selecting the best outcome from all threads, hence displaying the best performance.

\vspace{-1em}

  
  
  
\section{Conclusion}
\label{sec:conclusion}

We present a two-stage optimization workflow that combines dataless differentiable initialization with SOTA ILP solvers to address SDC-based combinatorial scheduling. By generating confident partial solutions through constrained Gumbel-Softmax sampling, our approach enables high-performance warm-start for various solvers, achieving 10$\times$ performance gain while achieving near-optimality ($<0.1\%$ gap). This work demonstrates that differentiable initialization and classical exact optimizers can synergistically overcome the speed-quality tradeoff inherent to SDC-formulated combinatorial scheduling problems. The workflow retains generality—requiring no labeled data and supporting GPU acceleration and combines exact-method optimality that inspires to application of similar strategies to broader combinatorial optimization problems.

\bibliographystyle{unsrt}
\bibliography{ref}

\end{document}